\documentclass[11pt,a4paper]{article}

\usepackage[frozencache,cachedir=.]{minted}
\usepackage[hyperref]{emnlp2020}
\usepackage{times}
\usepackage{latexsym}
\renewcommand{\UrlFont}{\ttfamily\small}

\usepackage{microtype}

\usepackage{makecell}
\usepackage{booktabs}
\usepackage{geometry}
\usepackage{graphicx}
\usepackage{caption}
\usepackage{xcolor}

\usepackage{array}
\usepackage{pifont}
\usepackage{tabularx}

\usepackage{tikz}
\usetikzlibrary{positioning}
\usepackage{subfig}

\definecolor{mygray}{RGB}{230, 230, 230}
\newcommand*\rot{\rotatebox{90}}
\newcommand*\OK{\ding{51}}

\aclfinalcopy 

\newcommand{\python}{\textsc{Python}}
\newcommand{\ourmodel}{\textsc{PyMT5}}

\title{\ourmodel: multi-mode translation of natural language and \python~ code with transformers
}

\author{Colin B. Clement\thanks{~~Corresponding author}\\
  Microsoft Cloud and AI\\
  \texttt{coclemen@microsoft.com}\And

  Dawn Drain \\
  Microsoft Cloud and AI\\
  \texttt{dadrain@microsoft.com}\AND
  
  Jonathan Timcheck\thanks{~~Work done during a Microsoft internship} \\
  Stanford University \\
  \texttt{timcheck@stanford.edu}\And
  
  Alexey Svyatkovskiy\\
  Microsoft Cloud and AI\\
  \texttt{alsvyatk@microsoft.com} \AND

  Neel Sundaresan \\
  Microsoft Cloud and AI\\
  \texttt{neels@microsoft.com}
  }

\date{}

\begin{document}
\maketitle
\begin{abstract}
Simultaneously modeling source code and natural language has many exciting applications in automated software development and understanding. Pursuant to achieving such technology, we introduce \ourmodel, the \python~method text-to-text transfer transformer, which is trained to translate between all pairs of \python~method feature combinations: a single model that can both predict whole methods from natural language documentation strings (docstrings) and summarize code into docstrings of any common style. We present an analysis and modeling effort of a large-scale parallel corpus of 26 million \python~methods and 7.7 million method-docstring pairs, demonstrating that for docstring and method generation, \ourmodel~ outperforms similarly-sized auto-regressive language models (\textsc{GPT2}) which were English pre-trained or randomly initialized. On the \textsc{CodeSearchNet} test set, our best model predicts 92.1\% syntactically correct method bodies, achieved a BLEU score of 8.59 for method generation and 16.3 for docstring generation (summarization), and achieved a ROUGE-L F-score of 24.8 for method generation and 36.7 for docstring generation.
\end{abstract}

\section{Introduction}

Software is a keystone of modern society, touching billions of people through services and devices daily. Writing and documenting the source code of this software are challenging and labor-intensive tasks; software developers need to repeatedly refer to online documentation resources in order to understand existing code bases to make progress. Developer productivity can be improved by the presence of source code documentation and a development environment featuring intelligent, machine-learning-based code completion and analysis tools.

Recent progress in natural language processing (NLP), especially encoder/decoder-based transformer models~\cite{vaswani2017attention} and pre-training~\cite{radford2018improving,lewis2019bart}, has led to state-of-the-art performance on language modeling, classification~\cite{devlin2018bert},  translation~\cite{raffel2019exploring}, summarization~\cite{liu2019text}, grammar correction~\cite{bryant2017automatic}, entity recognition, dialogue generation~\cite{budzianowski2019hello}, and more. Along with these quantitative advances have come deeper understanding of the learned hidden representations which power transformers~\cite{kovaleva2019revealing,voita2019bottom,clark2019does,ethayarajh2019contextual}. While they are arguably not `natural,' programming languages are increasingly becoming modeling playgrounds for NLP modeling. Since these languages by definition have a grammar, syntax, and known relationships between entities, they offer enticing opportunities for an even deeper probing of NLP models and tasks. Beyond theoretical importance, many NLP tasks have practical utility in software development environments: language modeling or generation can be used for code completion~\cite{raychev2014code, bruch2009learning,svyatkovskiy2019pythia,svyatkovskiy2020intellicode}, translation/summarization to generate documentation or natural language summaries~\cite{moreno2013automatic,scalabrino2017automatically,wan2018improving,alon2018code2seq} or even summarize a set of code changes~\cite{moreno2014automatic}, translation and grammar error correction to patch and detect bugs~\cite{zhai2019cpc}, and joint embedding of code and natural language for code search~\cite{husain2019codesearchnet, codeSearch2018}.

In this work we focus on jointly modeling both source code (\python) and concomitant natural language documentation (docstrings) with transformers, through the study of dual tasks: generating method code bodies from signatures and docstrings, and generating docstrings from signatures and method code bodies. While previous work~\cite{bimodalMiltos,yin-neubig-2017-syntactic} has leveraged the grammar of code to extract features like the Abstract Syntax Tree for modeling (treating code and natural language as separate modalities), we follow examples like~\citet{barone2017parallel} and treat \python~ and its docstrings as fundamentally no different than other `natural' languages, representing both source code and natural language docstrings as sequences of tokens sharing the same vocabulary. Here we present a multi-mode translation method resulting in \ourmodel, the \python~method text-to-text transfer transformer (inspired by the text-to-text transfer transformer T5~\cite{raffel2019exploring}). Our single model can both learn code/language generation and understand the relationships between them.

The paper is organized as follows: we begin in sec.~\ref{sec:multimode} by presenting examples of the performance of our novel multi-mode \ourmodel~---the \python~method text-to-text transfer transformer model---which we trained to translate between all pairs of combinations of method signatures, docstrings, and bodies which do not have the same feature in both the source and target. In sec.~\ref{sec:dataset} we describe our training data and the pre-processing steps for source code and natural language we followed, and compared it to existing parallel docstring-method corpora like \textsc{CodeSearchNet} (CSN)\cite{husain2019codesearchnet} and that presented by Barone et al~\cite{barone2017parallel}. In sec.\ref{sec:pretraining} we explain our BART-like~\cite{lewis2019bart} pre-training scheme, demonstrating a 25$\times$ speed-up in training time for docstring generation. Next, in sec.~\ref{sec:docstring-analysis} we analyze and classify \python~docstrings, enabling style-conditioned docstring generation in \ourmodel. In sections~\ref{sec:method-gen} and \ref{sec:docstring-gen}, we discuss \ourmodel~ results on method generation and docstring generation respectively and compare it to two GPT2 models randomly initialized and pre-trained on English.

\section{Multi-mode training}
\label{sec:multimode}

\begin{figure*}
    \centering
    \includegraphics[width=\textwidth]{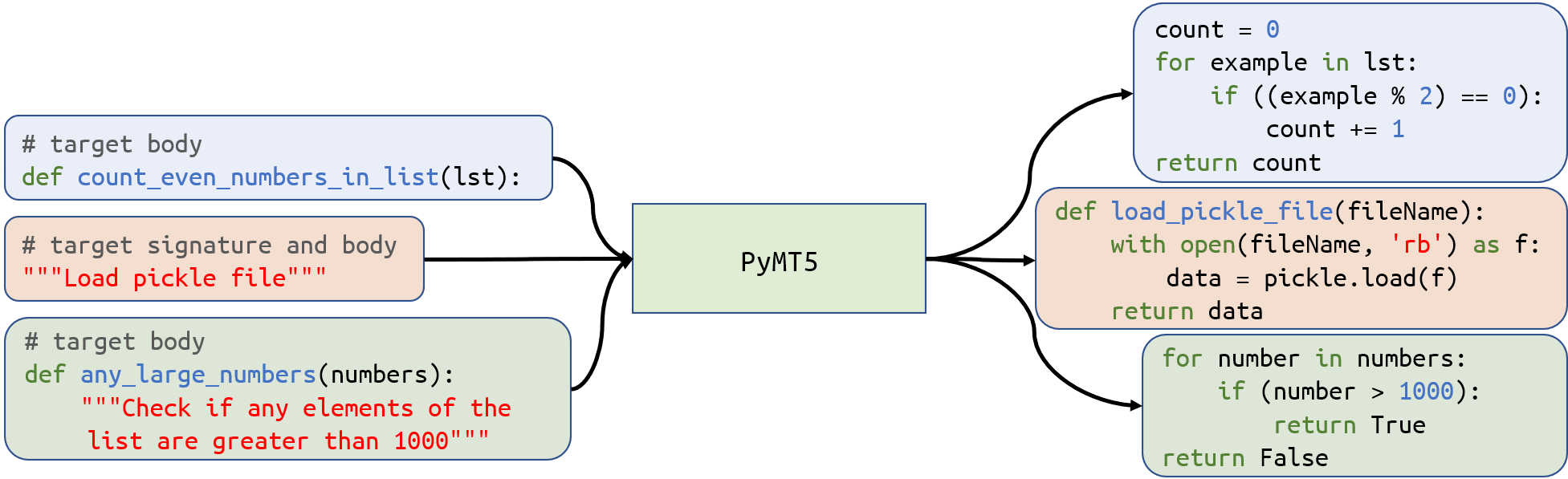}
    \caption{
    Real examples of \ourmodel~ performing method generation using combinations of signatures and docstrings. A leading comment in the input sequence instructs the model to output a particular combination of features, e.g. `\texttt{\# target signature and body}' instructs \ourmodel~ to predict both a signature and body.
    }
    \label{fig:multi-mode-schematic}
\end{figure*}

\tikzstyle{bluebox} = [draw=darkblue, fill=blue!6,
    rectangle, rounded corners] 
\tikzstyle{greenbox} = [draw=darkblue, fill=green!10,
    rectangle,  inner sep=5pt, inner ysep=5pt]    
\tikzstyle{redbox} = [draw=darkblue, fill=red!15,
    rectangle, rounded corners] 
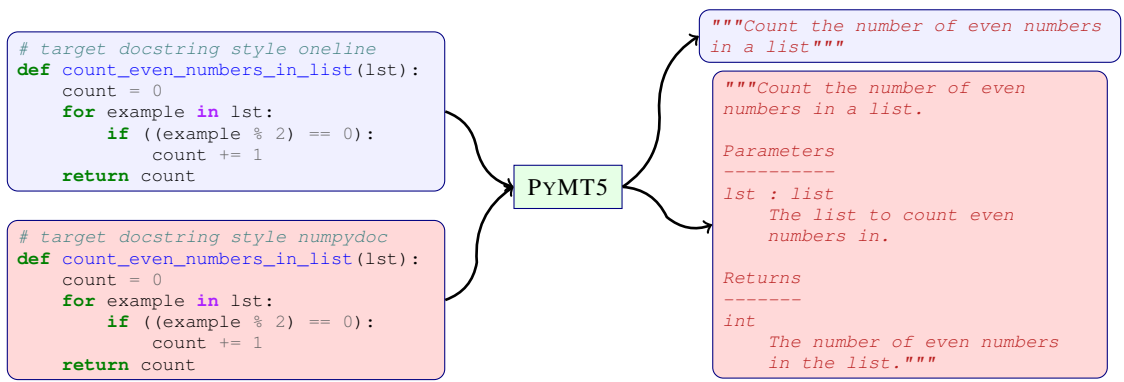
\begin{figure*}[tb]
\begin{center}
\begin{tikzpicture}[scale=1]
\node[bluebox] at (0, 2) (prompt-oneline) {%
\begin{minipage}{0.78\columnwidth}
\begin{minted}[fontsize=\scriptsize]{python}
# target docstring style oneline
def count_even_numbers_in_list(lst):
    count = 0
    for example in lst:
        if ((example % 2) == 0):
            count += 1
    return count
\end{minted}
\end{minipage}
};

\node[redbox] at (0, -.5) (prompt-numpydoc) {%
\begin{minipage}{0.78\columnwidth}
\begin{minted}[fontsize=\scriptsize]{python}
# target docstring style numpydoc
def count_even_numbers_in_list(lst):
    count = 0
    for example in lst:
        if ((example % 2) == 0):
            count += 1
    return count
\end{minted}
\end{minipage}
};

\node[greenbox] at (4.5, 1) (model) {\footnotesize \ourmodel};

\node[bluebox] at (9, 3.) (gen-oneline) {%
\begin{minipage}{0.75\columnwidth}\begin{minted}[fontsize=\scriptsize]{python}
"""Count the number of even numbers
in a list"""
\end{minted}
\end{minipage}
};

\node[redbox] at (9, 0.5) (gen-numpy) {%
\begin{minipage}{0.7\columnwidth}\begin{minted}[fontsize=\scriptsize]{python}
"""Count the number of even 
numbers in a list.

Parameters
----------
lst : list
    The list to count even 
    numbers in.

Returns
-------
int
    The number of even numbers 
    in the list."""
\end{minted}
\end{minipage}
};
\draw [line width=1pt, align=left, ->] (prompt-oneline.east) to [bend left] (3.3, 1.5) to [bend right] (model.west);
\draw [line width=1pt, align=left, ->] (prompt-numpydoc.east) to [bend right] (3.3, 0) to [bend left] (model.west);
\draw [line width=1pt, align=left, ->] (model.east) to [bend right] (5.7, 2.) to [bend left] (gen-oneline.west);
\draw [line width=1pt, align=left, ->] (model.east) to [bend left] (5.8, 0.6) to [bend right] (gen-numpy.west);
\end{tikzpicture}
    \caption{\ourmodel~ performing docstring generation on an example method, showing the output when the target prefix indicates one line (top, blue) and Numpydoc docstring (bottom, red) styles.} 
    \label{fig:generated_docstring}
\end{center}
\end{figure*}

Figure~\ref{fig:multi-mode-schematic} shows examples of inputs and outputs of our model \ourmodel~ for 3 example tasks: (top, blue) predicting a body from a method signature, (middle, red) predicting a whole method from a natural language docstring, and (bottom, green) predicting a body from a signature and docstring. Note that the comment `\texttt{\# target <specification>}' instructs the model to choose a particular form of output. Further note that \ourmodel~ correctly learns to interpret natural language: it interprets `\texttt{even}' as being related to `\texttt{(example \%2) == 0}', and `\texttt{greater than 1000}' as `\texttt{number > 1000}'. The model also produces syntactically correct code (as we will discuss later, we never show the model syntactically incorrect code), and correctly infers the types of `\texttt{lst}' and `\texttt{numbers}' to be iterables containing numbers.

\ourmodel~ can also be prompted with source code to produce a docstring summary in various styles. Figure~\ref{fig:generated_docstring} shows the model prompted with one of the methods generated by \ourmodel~ in Fig.~\ref{fig:multi-mode-schematic}~(top, blue), in both a `one line' (top, blue) style and a `Numpydoc' (bottom, red) style. It infers the intent from the signature name and code, and even infers that type of the argument is a \texttt{list} and return type \texttt{int}. It produces the same terse one sentence summary of the function in both cases.

In order to teach \ourmodel~ to maximally relate the separate method features (signatures, docstrings, bodies), we trained it to translate between all pairs of feature combinations in which the same feature does not appear in both the source and target. This scheme is also advantageous as our corpus is unbalanced, with only 1/5 methods featuring docstrings, and so the model can learn to leverage all the features whether they are present or not. Additionally, it has been shown that code is more `predictable' than natural language~\cite{hindle2012naturalness}. If the method and argument names are a dominating signal due to their relatively rigid structure, the model may learn to ignore the content of docstrings. This multi-mode method overcomes that by training the model to generate method bodies from docstrings alone. See the appendix for a more detailed description of the multi-mode training scheme.

\subsection{Dataset}
\label{sec:dataset}

\begin{table*}[htbp]
    \centering
    \small
    \begin{tabular}{l l l l}
        Dataset & Methods & w/ docstring & Languages\\
        \cmidrule{1-4}
        \ourmodel & $2.6\times 10^7$ & $7.7\times 10^6$ & \python~ \\
        CSN~\cite{husain2019codesearchnet} & $6.4\times 10^6$ & $2.3\times 10^6$ & \python, et al.\\
        \citet{ciurumelea2020suggesting} & $1.6\times 10^5$ & $1.6\times 10^5$ & \python\\
        \citet{barone2017parallel} & $1.6\times 10^5$ & $1.5\times 10^5$ & \python\\
    \end{tabular}
    \caption{
    Summary statistics of our \python~parallel corpus compared to others presented in the literature. CSN contains 500k \python~ methods with docstrings, among 6 other languages. Our parallel corpus is 3$\times$ as large as the next largest, and over 15$\times$ the size of the next largest \python~parallel corpus.
    }
    \label{tab:datastats}
\end{table*}

Our data consists of 118k \textsc{GitHub} repositories, which includes all public repositories labelled as containing primarily \python~source code, featuring at least 10 stars, and which have had a commit in the past 5 years. We successfully cloned 112k of these repositories, extracting 5.3 million \python~ files from the default \texttt{HEAD} state of each repository. We then removed literal duplicate files, resulting in 2.3 million unique files, but did not remove finer-grained clones. After removing license from the files, the literal contents were used in the pre-training step, comprising about 27\textsc{GB} of raw text.

In order to extract method-level information for fine-tuning, we used the \texttt{python3.7} standard library \texttt{ast} to produce the file-level Abstract Syntax Tree (AST) for each \python~file, extracting every individual and class method. For each file which failed to parse, we used \texttt{2to3} and \texttt{autopep8} to overcome the issue of different styles and white space or tab conventions, successfully parsing 97.3\% of the 2.3 million unique \python~ files.  We used the \python~module \texttt{astunparse} to take the AST for each method and unparse them back into source code, so that our fine-tuned model was never trained on syntactically incorrect code. The statistics of our method-docstring corpus are summarized in Table.~\ref{tab:datastats}. Our parallel method-docstring corpus is twice as large as the next largest irrespective of language and over 15$\times$ as large as the next largest \python~parallel corpus, both in CSN.

For each method, we ignored comments as they generally represent trivia and are not part of the normal language syntax. We cleaned the docstrings by removing non-\textsc{ASCII} characters, normalizing Unicode, and replacing commit hashes, file paths, and \textsc{URL}s with placeholder tokens. In all studies here, we randomly split the files at the repository level (to prevent data leakage) with 90\% for training, 5\% for validation, and 5\% for a test set.

\subsection{Pre-training}
\label{sec:pretraining}

\begin{figure*}[tb]
    \centering
    \includegraphics[width=0.8\textwidth]{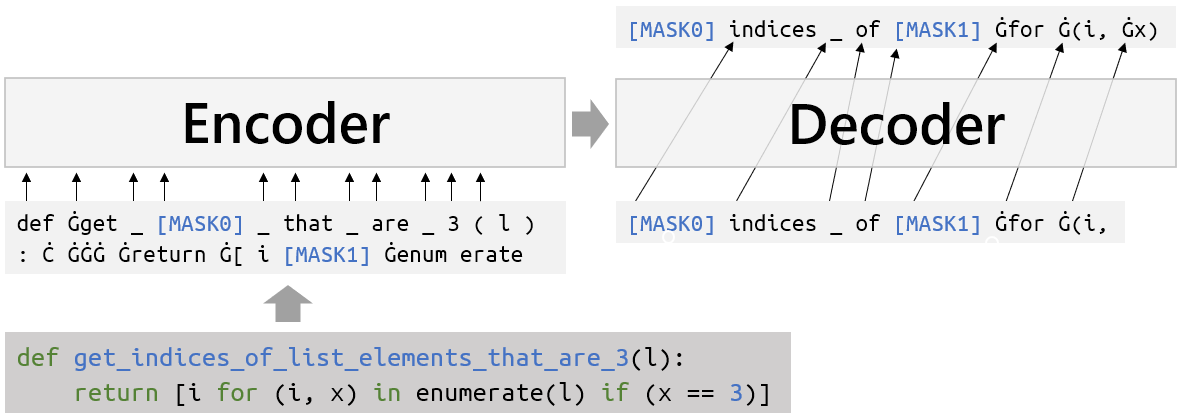}
    \caption{
    Denoising auto-encoder pre-training for sequence-to-sequence tasks, based on the span-masking objective used by the T5~\cite{raffel2019exploring}. \python~ files are first tokenized with spaces replaced by the character Ġ, which is 256 in ordinal above the space character (similarly for newlines, tabs, etc.). Note that indentation is a token of multiple Ġ's. We replace random sub-sequences of tokens with numbered masks, and train the model to return each mask followed by the tokens it replaced.
    }
    \label{fig:denoising}
\end{figure*}

The majority of our \python~methods---over 20 million methods--- do not possess docstrings. This imbalance is, in fact, an opportunity in light of the recent trend for NLP: unsupervised pre-training of language models on vast amounts of raw text~\cite{devlin2018bert}. Using these pre-trained models as starting points for downstream tasks---like classification, translation, summarization, and question answering---consistently yields state-of-the-art results~\cite{lewis2019bart,raffel2019exploring}.

Following this trend, we use a similar span-masking objective used by the recent text-to-text transfer transformer (T5)~\cite{raffel2019exploring}. As shown in Figure~\ref{fig:denoising}, after tokenizing the inputs, we sample a random subset of the token spans up to length 3 to be replaced with, e.g. a \texttt{[MASK0]} token, and then teach the sequence-to-sequence model to replace the missing tokens. The training target is comprised of numbered mask tokens followed by the tokens that mask represents.

The architecture of \ourmodel~ is an encode-decoder transformer with a vocabulary of 50181 (byte-pair BPE encoder trained on raw python files), 6 self-attention encoder/decoder layers in each encoder layers, and a hidden dimension of 1472, totaling 374 million parameters. All the experiments in this paper, including GPT2 were done using this same extended GPT tokenizer. We pre-trained \ourmodel~ on 27\textsc{GB} of raw source code in total, for 3 weeks on sixteen 32\textsc{GB} Tesla V100 \textsc{GPU}s, or 73 epochs total. When training on docstring generation alone, we observed 25$\times$ faster convergence to a lower loss when starting with this pre-trained model as compared to a random initialization. See the appendix for details. In all experiments \ourmodel~ is trained starting with this pre-trained model.

\subsection{Docstring analysis}
\label{sec:docstring-analysis}

When examining docstring samples from our corpus, one of the most salient features is the different styles of documentation. The \python~community has no prescribed or de facto style for docstrings, but \python~enhancement protocol 257~\cite{pep257} does describe one-line and multi-line docstrings, and mandates indentation as well. Most modern large-scale projects utilize docstring styles which are parseable, allowing the automatic creation and synchronization of source code and documentation websites, see, e.g. \texttt{sphinx}. Therefore, a number of standard styles have evolved in the community.

The currently dominant parseable docstring styles (and the ones supported by \texttt{sphinx}) are \textsc{reStructuredText} (reST)~\cite{jones2013restructuredtext}, the official \textsc{Google} style~\cite{google2020style}, \textsc{Numpy} style (also technically satisfies reST)~\cite{numpy2020style}, and \textsc{javadoc} style~\cite{javadoc2020style}. The difference between each style is mainly in the syntax of denoting sections (if they exist) and the name/type/description annotation of the method arguments and returned/yielded quantities (if they exist). We defined, in addition to these styles, one-line (containing only one line), one-paragraph (containing no empty lines), and `other' to label any docstring not described so far, which includes informal user docstring styles and a few project-specific styles like the \textsc{SAGE} mathematics toolkit library.

Table~\ref{tab:docstyle-stats} shows the breakdown of the fraction of each of these styles in our corpus. The plurality of docstrings (44\%) are one-line. The next most common style is one-paragraph at 14\%. The next four most-common styles are the machine parseable styles discussed above, comprising 26.2\% of the total number of docstrings. The appendix contains detailed distributions of method signature, docstring, and method body character and line lengths.

\begin{table}[htbp]
    \centering
    \small
    \begin{tabular}{l l}
        Style & Fraction of methods\\\hline
        One line & 44\% \\
        One paragraph & 14\% \\
        \textsc{reST} & 13\% \\
        \textsc{Google} & 7.3\% \\
        \textsc{Numpy} & 4.8\% \\
        \textsc{Javadoc} & 1.6\% \\
        Other & 15\%
    \end{tabular}
    \caption{
    Docstring style statistics from 7.7 million \python docstrings.
    }
    \label{tab:docstyle-stats}
\end{table}

\begin{figure}[htbp]
    \centering
    \includegraphics[width=\columnwidth]{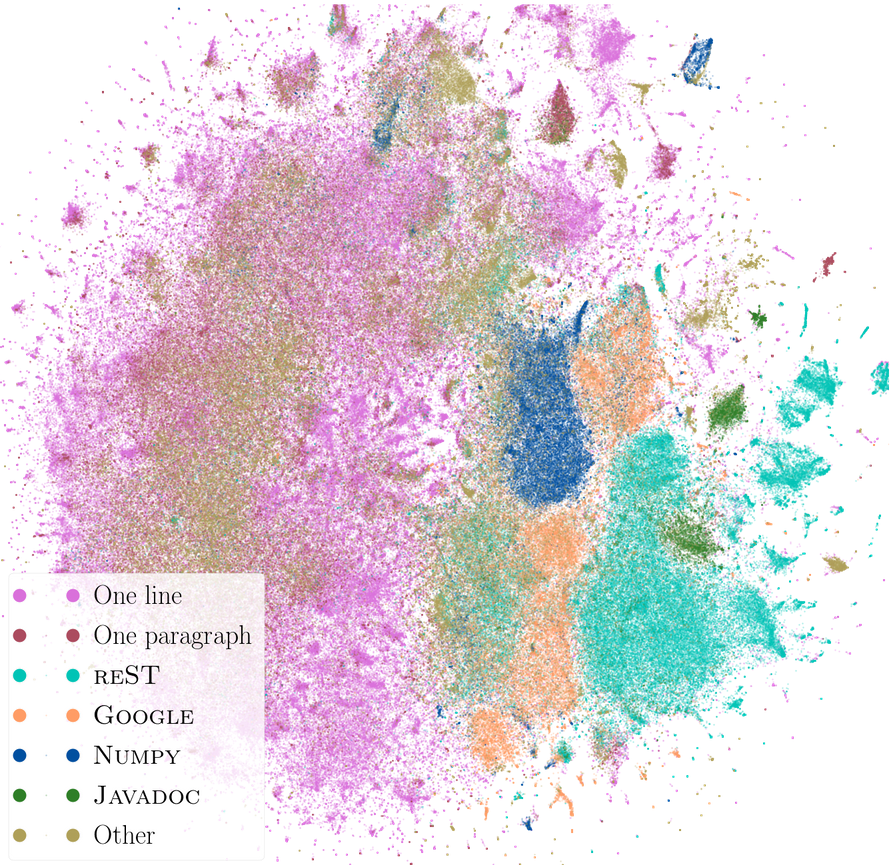}
    \caption{Visualization of continuous embeddings of 1/10th of our docstring corpus (770k docstrings), colored by docstring style. Embeddings were obtained using \textsc{FastText}, and the two-dimensional embedding was obtained via \textsc{PCA} (for dimensionality reduction and initialization) and t-\textsc{SNE}.}
    \label{fig:docstring-tsne}
\end{figure}

To visualize the space of these styles, we used \textsc{FastText} vector embeddings of the docstrings, obtaining 100-dimension continuous vector representations of each. We then used \textsc{PCA} to reduce the dimensionality to 50 and applied the t-distributed stochastic neighbor embedding (\textsc{t-sne}) to obtain a two-dimensional visualization. Figure~\ref{fig:docstring-tsne} shows 1/10th of our corpus (700k docstrings) embedded, colored by docstring style as defined above. We can see clear clustering of styles, indicating that similar docstrings use the same style (for the parseable styles). There is also a natural dichotomy between parseable and non-parseable styles: the left side is dominated by `one line,' `one paragraph,' and `other' styles, and the four parseable styles are largely on the right side. This observation can be used to generate documentation consistent with the style of a given project, or it could be used to translate methods into more informal descriptions useful for search indices.

\begin{table*}[tb]
    \centering
    \scriptsize
    \begin{tabular}{l l l l l l l l }
        Model & Ppl & BLEU & Syntax & Stat. & R1 & R2 & RL \\\hline\hline
        
        GPT2-med & 2.36 & 5.60 & 85\% &  Prec. & 25.8 & 12.3 & 26.8 \\
        random & & & &                             Rec.& 26.7 & 12.1 & 25.9 \\
        & & & &                             F1 & 21.8 & 10.6 & 22.5 \\\hline
        
        GPT2-med & \bf 2.09 & 5.63 & 86\% & Prec. & 25.4 & 12.1 & 26.3 \\
        english & & & &                             Rec. & 26.9 & 12.2 & 26.1 \\
        & & & &                             F1 & 21.7 & 10.6 & 22.5 \\\hline
        
        \ourmodel~ & 2.36 & {\bf 10.6} & \bf 93.6\% &  Prec. & \bf 33.8 & \bf 21.5 & \bf 33.6 \\
        & & & &                                         Rec. & \bf 44.1 & \bf 25.0 & \bf 43.8 \\
        & & & &                                         F1 & \bf 35.1 &  \bf 21.5 & \bf 32.2 \\\hline
        
        CSN test: &  & & & & & & \\\hline
        
        GPT2-med & -- & 2.8 & 77.2\% &  Prec. & \bf 32.3 & 11.8 & \bf 33.7 \\
        random & & & &                  Rec. & 19.6 & 7.0 & 19.3 \\
        & & & &                         F1 & 20.9 & 7.6 & 21.9 \\\hline        
        \ourmodel & -- & \bf 8.59 & \bf 92.1\% &   Prec. & 25.6 & \bf 12.5 & 25.3 \\
        & & & &                                     Rec. & \bf 40.2 & \bf 18.3 & \bf 39.6 \\
        & & & &                                     F1 & \bf 28.4 &  \bf 13.5 & \bf 24.8 \\\hline
        
        \citet{barone2017parallel} test: &  & & & & & & \\\hline
        
        \ourmodel & -- & 20.2 & 91.1\% &    Prec. & 41.3 & 28.5 & 40.7 \\
        & & & &                             Rec. & 52.2 & 34.7 & 51.3 \\
        & & & &                             F1 & 43.2 &  29.8 & 39.7 \\
        
        Barone et al. & -- & 10.9  &  --  & -- & -- & --      
    \end{tabular}
    \caption{
Comparing 3 models--\textsc{GPT2} with a random weight initialization, \textsc{GPT2} pre-trained on English, and \ourmodel--on the task of method generation from a signature and natural language docstring. The first three rows use our test set consisting of 1,285,794 methods. The fourth and fifth rows compare the performance of \ourmodel~ and GPT2-medium on the CodeSearchNet \python~test set. The final rows compare the performance of \ourmodel~on the parallel corpus test set of \citet{barone2017parallel}. Syntax is the fraction of predicted methods which had correct syntax using the \python~3.7 grammar. 
    }
    \label{tab:method-gen}
\end{table*}

\begin{table}[bt]
    \centering
    \scriptsize
    \begin{tabularx}{\columnwidth}{@{}l X X X l l l @{}}
        Model & Ppl & BLEU &   & R1 & R2 & RL \\\hline\hline
        
        GPT2-med & 2.36 & 19.4  &    P & 32.6 & 19.3 & 33.6 \\
        random & & &             R & 36.2 & 19.4 & 34.7 \\
        & & &                    F1 & 31.0 & 18.2 & 31.6 \\\hline
        
        GPT2-med & \bf 2.15 & 19.6 &   P & 33.1 & 19.4 & 33.9 \\
        English & & &           R & 36.4 & 19.5 & 34.8 \\
        & & &                   F1 & 31.4 & 18.3 & 31.8 \\\hline
        
        \ourmodel & 3.74 & {\bf 25.2}  &    P & \bf 42.1 & \bf 23.7 & \bf 41.3 \\
        & & &                               R & \bf 50.4 & \bf 27.0 & \bf 49.3 \\
        & & &                               F1 & \bf 43.3 &  \bf 24.4 & \bf 39.8 \\\hline
        
        CSN test: & & & & & & \\\hline

        
        GPT2-med & -- & 9.5  &      P & 30.6 & 13.3 & 31.4 \\
        random & & &                R & 31.1 & 12.9 & 29.8 \\
        & & &                       F1 & 26.3 &  11.5 & 27.2 \\\hline
        
        \ourmodel & -- & \bf 16.3  &    P & \bf 38.0 & \bf 19.2 & \bf 36.8 \\
        & & &                           R & \bf 52.7 & \bf 24.5 & \bf 51.0 \\
        & & &                           F1 & \bf 41.3 & \bf 20.4 & \bf 36.7 \\\hline      
        
        Barone test: & & & & & & \\\hline
        
        \ourmodel & -- & 17.4  &    P & 39.6 & 26.0 & 38.7 \\
        & & &                       R & 53.6 & 33.7 & 52.1 \\
        & & &                       F1 & 43.1 &  27.8 & 39.1 \\
        Barone et al. & -- & 13.84  &  --  & -- & -- & --
    \end{tabularx}
    \caption{
Comparing 3 models--\textsc{GPT2} with a random weight initialization, \textsc{GPT2} pre-trained on English, and \ourmodel--on the task of natural language docstring generation from a signature and method body. The first three rows are evaluated on our test set of 383695 methods. The fourth and fifth rows shows performance of \ourmodel~ and GPT2-medium on the CSN \python~test set, and the last two rows compare our model to Barone et al. on their test set.
    }
    \label{tab:docstring-gen}
\end{table}

\section{Method generation}
\label{sec:method-gen}

Now we turn our attention to method generation: predicting a whole method code body from either a method signature, a natural language docstring, or both. We first discuss a benchmark of this task using a GPT2-medium model (345 million parameters, see the appendix for details), training from scratch and starting with the publicly released \textsc{OpenAI} English pre-trained checkpoint with weights from HuggingFace\cite{Wolf2019HuggingFacesTS}. In all experiments we used an extended GPT2 tokenizer---including white-space (one tab, two tabs, etc.) tokens---for a total vocabulary size of 50337, and we used beam decoding with a beam width of 5.

The third row of tab.~\ref{tab:method-gen} shows \ourmodel~ has more than double the BLEU score, overall better recall, and significantly better ROUGE-2 and ROUGE-L F-scores than our GPT2 baselines. Further, 93.6\% of the methods generated by \ourmodel~ were syntactically correct \python~3.7, whereas only 86\% of GPT2 methods were syntactically correct. \ourmodel~was trained on 16 Tesla V100 16GB GPUs for 62 epochs, or 5 weeks training time (see the appendix for its hyper-parameters) and the GPT2 baselines were trained on the same hardware for 1 week training time (achieving the same or better validation loss/perplexity as \ourmodel).

The English pre-trained initialization of GPT2 only slightly beats the random initialization of GPT2, which could indicate that the learned biases of English are not particularly beneficial for writing \python~ code; the metrics are almost all within our margin of error. Note that \citet{barone2017parallel} also modeled methods from docstrings, obtaining a similar BLEU score of 10.9 on their own \python~ parallel corpus. On the Barone et al. test set, \ourmodel~obtains nearly double these scores at 20.2; such a large discrepancy could be explained by data leaking from their test set into our training set. Barone's test set is also 200$\times$ smaller than ours and may not be a representative sample of the whole \python~ code domain.

The third and fourth rows of tab.~\ref{tab:method-gen} show the performance of \ourmodel~ using the publicly available CSN \python~test set, from which we find notably worse results than on our own test set. CSN curated their whole set by removing any methods with `test' in the name and any methods with fewer than 3 lines of code. We calculated the performance of \ourmodel~only on a subset of our test set curated the same way as CSN, observing F-scores for R1, R2, and R-L on our test set of 29.7, 17.2, and 26.1, which is lower than our nominal test set performance of 35.1, 21.5, and 32.2 and closer to the CSN performance of 28.4, 13.5, and 24.8. We believe this curating choice explains the difference between our test set and the CSN test set. We also conclude that tests and short methods are `easier' to complete, which is plausible, and bodes well for automatic code completion applications.

\section{Docstring Generation}
\label{sec:docstring-gen}

We now examine results from the docstring generation task, which for evaluation purposes were conditioned on both signatures and method bodies. As in method generation, we set a GPT2 benchmark with random initialization and pre-trained English initialization as well as the same hyperparameters. Table~\ref{tab:docstring-gen} shows that the ROUGE scores of the GPT2 baselines are within the margin of error; a somewhat surprising result given the English domain of docstrings. The third row shows \ourmodel~ to be superior to GPT2-medium in terms of BLEU and all of the ROUGE metrics.

We again present the results from the publicly available CSN test set. Similar to the method generation task, \ourmodel~performs worse on the CSN data than our own, likely for the same reasons we discussed in sec.~\ref{sec:method-gen}. We also evaluated \ourmodel~ on the Barone et al. parallel test set, as shown in the second to last row of tab.~\ref{tab:docstring-gen}, and find \ourmodel~performs notably worse on Barone's test set than our own test set, contradicting the hypothesis that our doubling of the method generation BLEU score is due to data leakage. \ourmodel~has a much higher BLEU score than that reported by Barone et al, perhaps indicating real progress in the code summarization field.

Docstring generation is similar to code summarization, though the domains are different as docstrings also contain structured annotations of arguments, return values, raised exceptions, and even in-line unit tests (\texttt{doctest}). TranS$^3$ by Wang et al.~\cite{wang2020trans} reports a best ROUGE-L of 51.27 on the same test set for code summarization, but does not specify which statistic they are reporting, so we cannot make strong conclusions about the performance of \ourmodel~compared to the state of the art.

\section{Conclusion}

In this work, we presented a novel multi-mode \python~method text-to-text transfer transformer model \ourmodel as well as the largest parallel corpus of \python~source code and docstrings reported in the literature to date. We have trained \ourmodel~ to translate between all pairs of combinations of method signatures, docstrings, and method bodies which do not have the same feature in both the source and target. Further, we introduced control token prefixes for docstring generation to facilitate docstring generation of various styles. Focusing on two modeling tasks -- predicting \python~methods from docstrings and summarizing \python~source code methods into docstrings of various commonly occurring styles -- we have compared this new approach to the auto-regressive \textsc{GPT2} baselines trained on individual docstring or method generation tasks. On the \textsc{CodeSearchNet} test set \ourmodel~ achieves a BLEU score of 8.59 for method generation and 16.3 for docstring generation, and a ROUGE-L F-score of 24.8 for method generation and 36.7 for docstring generation. We have demonstrated the effectiveness of dynamic masked pre-training, reducing docstring generation training time by 25$\times$. Looking forward, we plan to leverage \ourmodel~ for various downstream automated software engineering tasks---including code documentation and method generation from natural language statements---and develop more model evaluation criteria to leverage the unique properties of source codes.

\section*{Acknowledgements}

We would like to thank the Microsoft Cloud and AI SmartML engineering team for help in preparing the data, Shao Kun Deng for the development of compelling user experiences leveraging \ourmodel, and Christian Bird for useful discussions.

\newcommand{\appendixdefguard}{}
\ifdefined\appendixdefguard

\else

\documentclass[11pt,a4paper]{article}
\usepackage[hyperref]{emnlp2020}
\usepackage{times}
\usepackage{latexsym}
\renewcommand{\UrlFont}{\ttfamily\small}

\usepackage{microtype}
\usepackage{makecell}
\usepackage{booktabs}
\usepackage{geometry}
\usepackage{graphicx}
\usepackage{caption}
\usepackage{xcolor}
\usepackage{minted}
\usepackage{array}
\usepackage{pifont}

\usepackage{tikz}
\usetikzlibrary{positioning}
\usepackage{subfig}

\title{\ourmodel: multi-mode translation of natural language and \python~ code with transformers}

\begin{document}

\definecolor{mygray}{RGB}{230, 230, 230}
\newcommand*\rot{\rotatebox{90}}
\newcommand*\OK{\ding{51}}

\newcommand{\python}{\texttt{python}}
\newcommand{\ourmodel}{\textsc{PyMT5}}

\fi

\aclfinalcopy 



\maketitle

\appendix

\section{Appendix}
\label{sec:appendix}

\subsection{Docstring statistics}
\begin{figure*}[tb]
    \centering
    \includegraphics[width=0.85\textwidth]{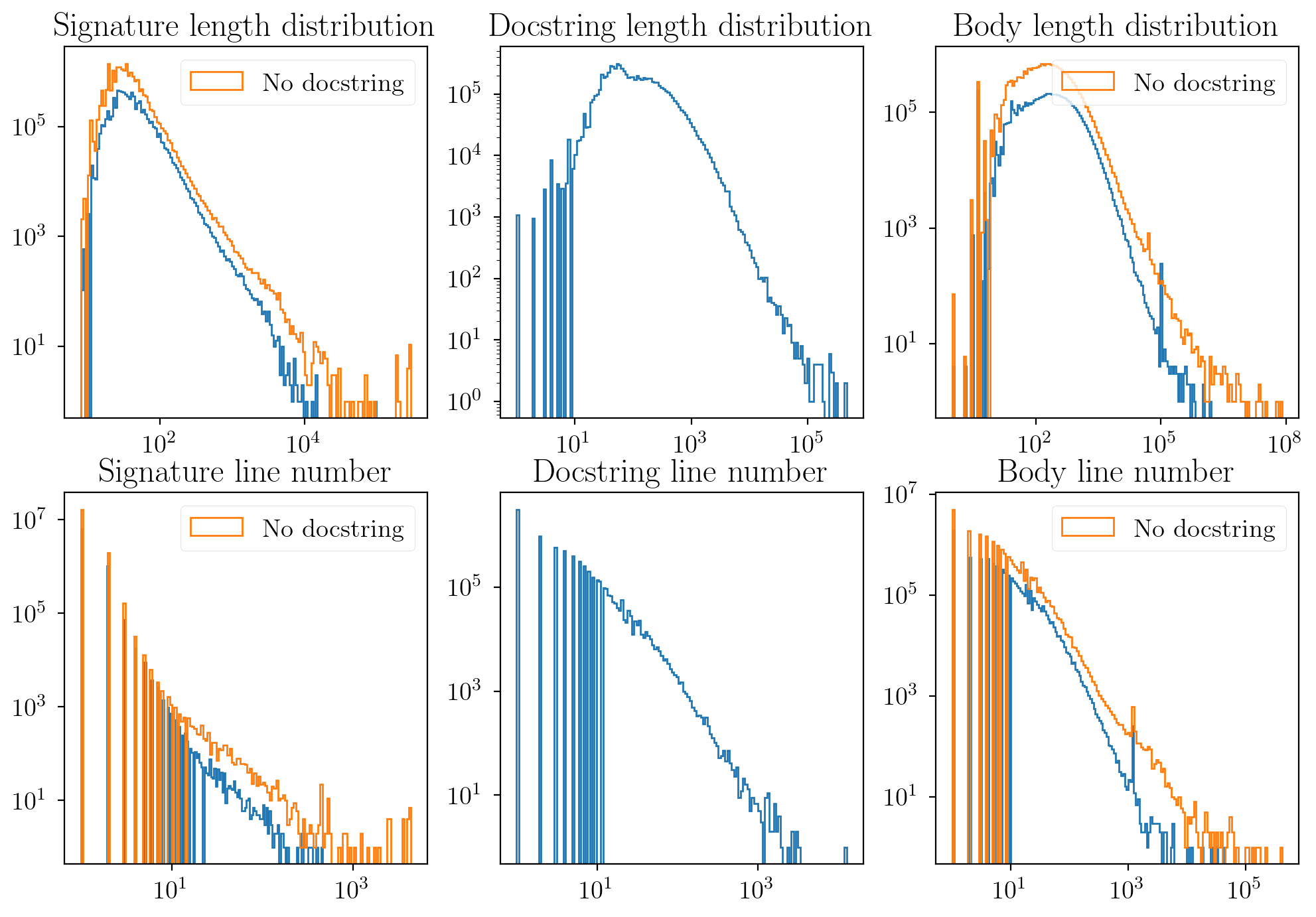}
    \caption{
    Histogram of the number of characters (top row) in the \python~signatures (left), docstrings (middle), and method body (right). The blue lines are for methods with docstrings, the yellow lines are for methods without docstrings. The vast majority of docstrings have more than 10 characters. The bottom row shows histograms of the number of lines for the same features described in the top row.
    }
    \label{fig:method-distn}
\end{figure*}

Figure~\ref{fig:method-distn} shows the distributions of various features of docstrings in our corpus. The top row is the distribution of total character-level length of the method signatures (left), docstrings (center), and code bodies. The blue lines are for methods possessing a docstring, and we can see that the vast majority of these methods have docstrings with more than 10 characters. The bottom row shows the distribution of line lengths of the concomitant features from the top row. While the most common line length of docstrings is 1 (comprising 41\%), the vast majority of docstrings have multiple lines.

\subsection{Pre-training details}

Figure~\ref{fig:pre-train-script} is the complete training script, using the Facebook AI Research Sequence (\textsc{FairSeq}) modeling library, with which we pre-trained \ourmodel. The data was pre-noised and processed using the \texttt{fairseq-preprocess} command, and placed in the directory indicated by \texttt{\$DIR}. The architecture and training hyper-parameters are set in this script. \ourmodel~ was trained with the same hyperparameters, but with data described in sec.\ref{sec:multi-mode-appendix}.

Figure~\ref{fig:pre-train-script} shows learning curves of a single seq2seq model of the same architecture as \ourmodel~ trained only on docstrings, starting from random initializations, and starting from our pre-trained model. As the figure shows, the pre-trained initialization converged to a better validation loss 25$\times$ faster than the randomly initialized model.

\begin{figure}
    \centering
    \includegraphics[width=\columnwidth]{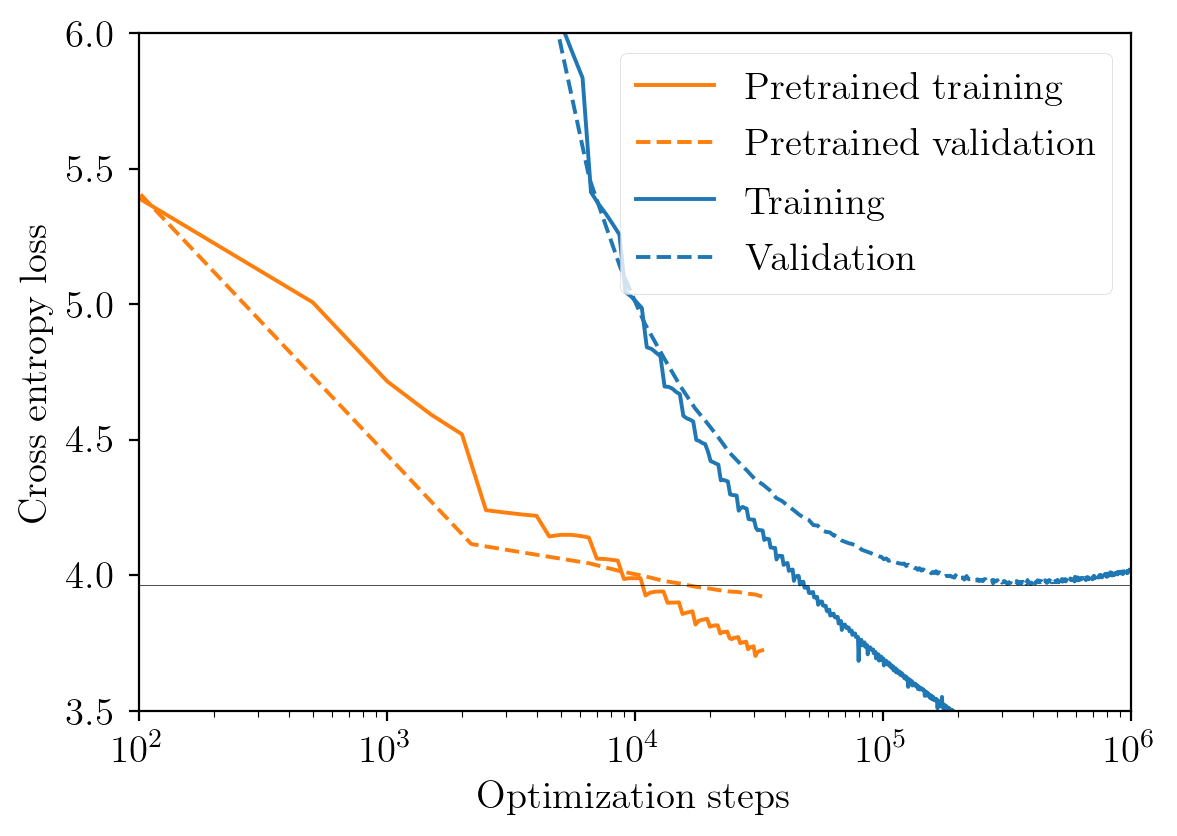}
    \caption{
    Learning curves for training a sequence-to-sequence transformer, translating from python method definitions to their docstrings. Blue curves represent the training and validation loss, and show that convergence (validation loss stops decreasing) occurs after $3.97\times 10^5$ steps or 183 epochs. The optimization of the pre-trained model with identical hyperparameters reaches and beats the best validation loss at $1.5\times 10^4$ steps or 7 epochs.
    }
    \label{fig:pretrain-learning-curve}
\end{figure}

\begin{figure}[tb]
\begin{minted}[fontsize=\scriptsize]{bash}
TOTAL_NUM_UPDATES=1300000
WARMUP_UPDATES=5000
LR=9.1875e-05
MAX_TOKENS=2200
UPDATE_FREQ=64
DIR=<data-dir>

fairseq-train $DIR$ \
    --max-tokens $MAX_TOKENS \
    --task translation \
    --source-lang src --target-lang tgt \
    --share-all-embeddings \
    --share-decoder-input-output-embed \
    --arch transformer \
    --dropout 0.2 --relu-dropout 0.2 \
    --attention-dropout 0.2 \
    --encoder-embed-dim 1472 \
    --decoder-embed-dim 1472 \
    --max-target-positions 1024 \
    --max-source-positions 1024 \
    --encoder-ffn-embed-dim 4096 \
    --decoder-ffn-embed-dim 4096 \
    --encoder-attention-heads 8 \
    --decoder-attention-heads 8 \
    --criterion label_smoothed_cross_entropy \
    --label-smoothing 0.1 \
    --dropout 0.1 --attention-dropout 0.1 \
    --weight-decay 0.01 --optimizer adam \
    --clip-norm 0.1 \
    --lr-scheduler inverse_sqrt --lr $LR \
    --warmup-updates $WARMUP_UPDATES \
    --update-freq $UPDATE_FREQ \
    --skip-invalid-size-inputs-valid-test \
    --save-dir $DIR/models \
    --save-interval 16 \
    --fp16 --adam-betas '(0.9,0.98)' \
    --adam-eps 1e-6 \
    --tensorboard-logdir $DIR/tensorboard \
    --decoder-learned-pos --encoder-learned-pos
  \end{minted}
  \caption{The \texttt{fairseq-train} script used to  pre-train \ourmodel, setting all the relevant hyper-parameters.}
  \label{fig:pre-train-script}
\end{figure}

\subsection{GPT2 training details}

Our GPT2 experiments also used the \textsc{FairSeq} library, with the OpenAI English checkpoint supplied by the HuggingFace library. Figure~\ref{fig:gpt-training-script} shows the complete training script, where for the English pre-trained initialization a pre-trained checkpoint was provided. Each models was trained on 4 Tesla V100 GPUs with 16GB of memory each, for 7 days.

\begin{figure}[tb]
\begin{minted}[fontsize=\scriptsize]{bash}
fairseq-train $DIR \
  --task language_modeling \
  --optimizer adam \
  --adam-betas "(0.9, 0.98)" \
  --weight-decay 0.01 \
  --clip-norm 0.0 \
  --lr  0.0005 \
  --reset-optimizer \
  --lr-scheduler inverse_sqrt \
  --warmup-updates 4000 \
  --warmup-init-lr 1e-07 \
  --dropout 0.1 \
  --weight-decay 0.01 \
  --tokens-per-sample 1024 \
  --sample-break-mode complete \
  --max-tokens 4096 \
  --update-freq 4 \
  --fp16 \
  --arch hf_gpt2_medium \
  --max-target-positions 1024 \
  --skip-invalid-size-inputs-valid-test
  \end{minted}
    \caption{The \texttt{fairseq-train} script we used to train our GPT model baselines}
    \label{fig:gpt-training-script}
\end{figure}

\subsection{Multi-mode training details}
\label{sec:multi-mode-appendix}

In order to better teach \ourmodel~ to understand the relationships between all the different features of code (signatures, docstrings, and bodies) we taught it to translate between all pairs of combinations of these features which do not contain the same feature in both the source and target. In this way, the model can learn to produce method bodies using both signatures and docstrings, or one or the other. Table~\ref{tab:multimode} spells out exactly which combinations were provided to the model as a source and target. For each source example the comment string `\texttt{\# target <feature> (<style>)}' was added, instructing the model which feature combination (e.g. signature and body). Only if a docstring was in the target, a style imperative was added, where the styles are defined and discussed in the main text.

\begin{figure*}[tb]
    \centering
    \includegraphics[width=0.85\textwidth]{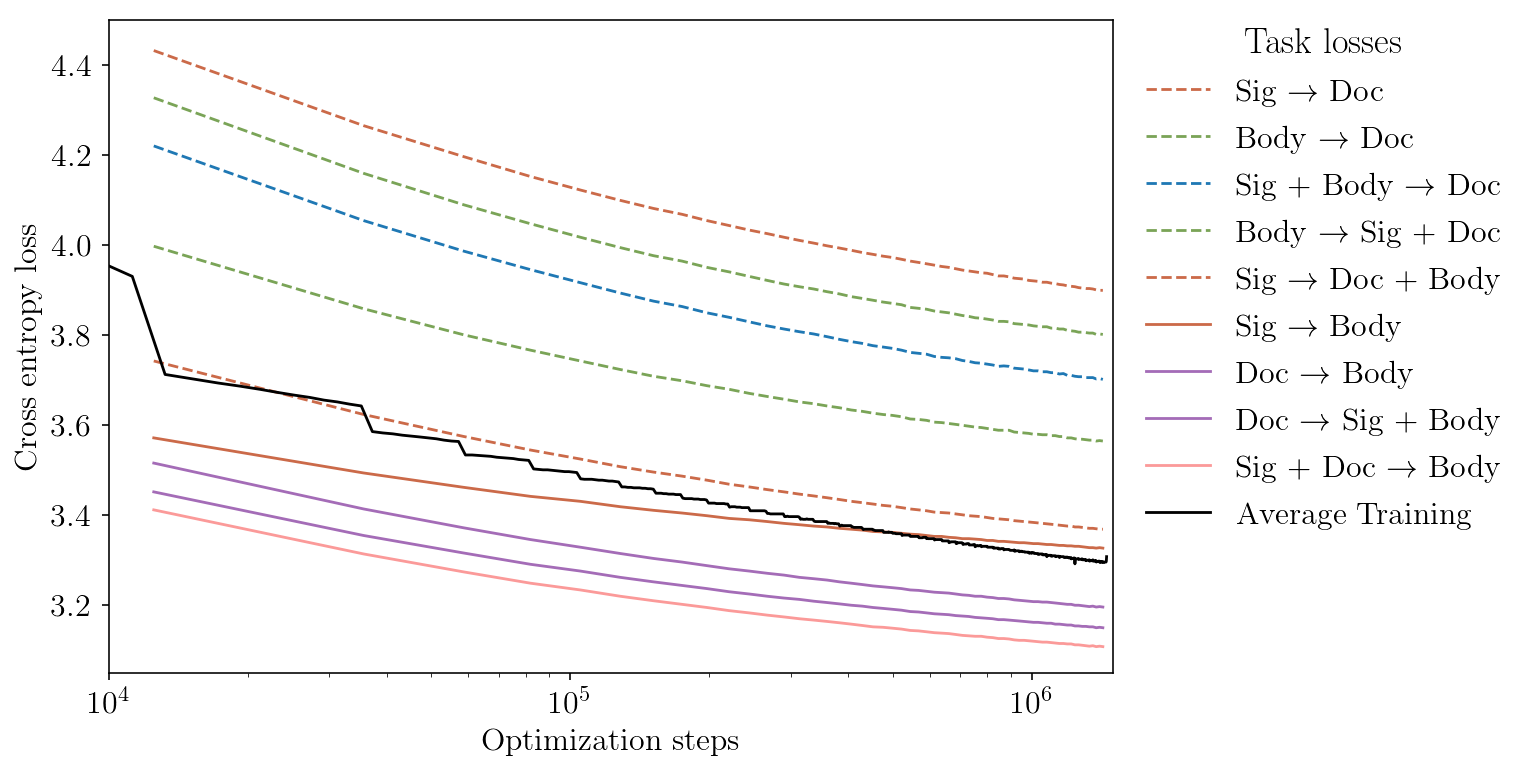}
    \caption{Learning curve for the multi-mode training, where the black line is the training loss, and the other lines are the validation loss for each mode of translation. Dashed lines indicate the docstrings are in the target, solid lines have only code in the target.
    }
    \label{fig:multi-mode}
\end{figure*}

Figure~\ref{fig:multi-mode} shows the training curves for \ourmodel, where the solid black line is the training loss, and all the other curves are the validation loss for each of the tasks indicated in tab.~\ref{tab:multimode}. The dashed lines indicate tasks where docstrings are present in the target, showing that these are generally less predictable than code-only targets (as the validation loss is larger). \ourmodel was trained on 16 Tesla V100 16GB GPUs for 62 epochs, or 5 weeks training time.

\begin{table}[tb]
\centering\small
    \begin{minipage}{\columnwidth}
    \begin{tabular}{ cl*{7}c }
        & & \multicolumn{7}{c}{Sources} \\
        & & \rot{Signature} & \rot{Dosctring} & \rot{Body} & \rot{Sig + doc} 
        & \rot{Sig + body} & \rot{Doc + body}  \\
        \cmidrule{2-8}
        & Signature         &      & \OK & \OK &     &     & \OK \\
        & Docstring         &  \OK &     & \OK &     & \OK &     \\
        \cmidrule{2-6}
        & Body              &  \OK & \OK &     & \OK &     &     \\
        \cmidrule{2-6}
        & Sig + doc         &      &     & \OK &     &     &     \\
                 \rot{~~~~~Targets}
        & Sig + body        &      & \OK &     &     &     &     \\
        & Doc + body        &  \OK &     &     &     &     &     \\
        \cmidrule[1pt]{2-8}
    \end{tabular}
    \end{minipage}
    \caption{A table of all possible translation possibilities between the 3 features of a function: the signature (sig), docstring (doc), and body. We train our model to translate between sources and targets indicated with a \OK, which were chosen as all pairs of feature combinations which do not contain the same feature in both the source and target. The system is then instructed to target code bodies when performing function completion.
    }
    \label{tab:multimode}

\end{table}

\ifdefined\appendixdefguard

\else
\end{document}
\fi

\bibliographystyle{acl_natbib}
\bibliography{emnlp2020}

\begin{thebibliography}{36}
\expandafter\ifx\csname natexlab\endcsname\relax\def\natexlab#1{#1}\fi

\bibitem[{jav(2011)}]{javadoc2020style}
 2011.
\newblock \href {http://wiki.c2.com/?JavaDoc} {Java doc}.
\newblock Technical report.

\bibitem[{Allamanis et~al.(2015)Allamanis, Tarlow, Gordon, and
  Wei}]{bimodalMiltos}
Miltiadis Allamanis, Daniel Tarlow, Andrew~D. Gordon, and Yi~Wei. 2015.
\newblock Bimodal modelling of source code and natural language.
\newblock In \emph{Proceedings of the 32nd International Conference on
  International Conference on Machine Learning - Volume 37}, ICML’15, page
  2123–2132. JMLR.org.

\bibitem[{Alon et~al.(2018)Alon, Brody, Levy, and Yahav}]{alon2018code2seq}
Uri Alon, Shaked Brody, Omer Levy, and Eran Yahav. 2018.
\newblock code2seq: Generating sequences from structured representations of
  code.
\newblock \emph{arXiv preprint arXiv:1808.01400}.

\bibitem[{Barone and Sennrich(2017)}]{barone2017parallel}
Antonio Valerio~Miceli Barone and Rico Sennrich. 2017.
\newblock A parallel corpus of python functions and documentation strings for
  automated code documentation and code generation.
\newblock \emph{arXiv preprint arXiv:1707.02275}.

\bibitem[{Bruch et~al.(2009)Bruch, Monperrus, and Mezini}]{bruch2009learning}
Marcel Bruch, Martin Monperrus, and Mira Mezini. 2009.
\newblock Learning from examples to improve code completion systems.
\newblock In \emph{Proceedings of the 7th joint meeting of the European
  software engineering conference and the ACM SIGSOFT symposium on the
  foundations of software engineering}, pages 213--222.

\bibitem[{Bryant et~al.(2017)Bryant, Felice, and Briscoe}]{bryant2017automatic}
Christopher Bryant, Mariano Felice, and Edward Briscoe. 2017.
\newblock Automatic annotation and evaluation of error types for grammatical
  error correction.
\newblock Association for Computational Linguistics.

\bibitem[{Budzianowski and Vuli{\'c}(2019)}]{budzianowski2019hello}
Pawe{\l} Budzianowski and Ivan Vuli{\'c}. 2019.
\newblock Hello, it's gpt-2--how can i help you? towards the use of pretrained
  language models for task-oriented dialogue systems.
\newblock \emph{arXiv preprint arXiv:1907.05774}.

\bibitem[{Ciurumelea et~al.(2020)Ciurumelea, Proksch, and
  Gall}]{ciurumelea2020suggesting}
Adelina Ciurumelea, Sebastian Proksch, and Harald Gall. 2020.
\newblock Suggesting comment completions for python using neural language
  models.
\newblock In \emph{27th edition of the IEEE International Conference on
  Software Analysis, Evolution and Reengineering (SANER)}. IEEE.

\bibitem[{Clark et~al.(2019)Clark, Khandelwal, Levy, and
  Manning}]{clark2019does}
Kevin Clark, Urvashi Khandelwal, Omer Levy, and Christopher~D Manning. 2019.
\newblock What does bert look at? an analysis of bert's attention.
\newblock \emph{arXiv preprint arXiv:1906.04341}.

\bibitem[{Devlin et~al.(2018)Devlin, Chang, Lee, and
  Toutanova}]{devlin2018bert}
Jacob Devlin, Ming-Wei Chang, Kenton Lee, and Kristina Toutanova. 2018.
\newblock Bert: Pre-training of deep bidirectional transformers for language
  understanding.
\newblock \emph{arXiv preprint arXiv:1810.04805}.

\bibitem[{Ethayarajh(2019)}]{ethayarajh2019contextual}
Kawin Ethayarajh. 2019.
\newblock How contextual are contextualized word representations? comparing the
  geometry of bert, elmo, and gpt-2 embeddings.
\newblock \emph{arXiv preprint arXiv:1909.00512}.

\bibitem[{Goodger and van Rossum(2001)}]{pep257}
David Goodger and Guido van Rossum. 2001.
\newblock \href {https://www.python.org/dev/peps/pep-0257/} {Docstring
  conventions}.
\newblock PEP 257.

\bibitem[{Google(2020)}]{google2020style}
Google. 2020.
\newblock \href {http://google.github.io/styleguide/pyguide.html} {Google
  python style guide}.
\newblock Technical report.

\bibitem[{Gu et~al.(2018)Gu, Zhang, and Kim}]{codeSearch2018}
Xiaodong Gu, Hongyu Zhang, and Sunghun Kim. 2018.
\newblock \href {https://doi.org/10.1145/3180155.3180167} {Deep code search}.
\newblock In \emph{Proceedings of the 40th International Conference on Software
  Engineering}, ICSE ’18, page 933–944, New York, NY, USA. Association for
  Computing Machinery.

\bibitem[{Hindle et~al.(2012)Hindle, Barr, Su, Gabel, and
  Devanbu}]{hindle2012naturalness}
Abram Hindle, Earl~T Barr, Zhendong Su, Mark Gabel, and Premkumar Devanbu.
  2012.
\newblock On the naturalness of software.
\newblock In \emph{2012 34th International Conference on Software Engineering
  (ICSE)}, pages 837--847. IEEE.

\bibitem[{Husain et~al.(2019)Husain, Wu, Gazit, Allamanis, and
  Brockschmidt}]{husain2019codesearchnet}
Hamel Husain, Ho-Hsiang Wu, Tiferet Gazit, Miltiadis Allamanis, and Marc
  Brockschmidt. 2019.
\newblock Codesearchnet challenge: Evaluating the state of semantic code
  search.
\newblock \emph{arXiv preprint arXiv:1909.09436}.

\bibitem[{Jones(2013)}]{jones2013restructuredtext}
Richard Jones. 2013.
\newblock A restructuredtext primer.
\newblock \emph{docutils. sourceforge. net, March}.

\bibitem[{Kovaleva et~al.(2019)Kovaleva, Romanov, Rogers, and
  Rumshisky}]{kovaleva2019revealing}
Olga Kovaleva, Alexey Romanov, Anna Rogers, and Anna Rumshisky. 2019.
\newblock Revealing the dark secrets of bert.
\newblock \emph{arXiv preprint arXiv:1908.08593}.

\bibitem[{Lewis et~al.(2019)Lewis, Liu, Goyal, Ghazvininejad, Mohamed, Levy,
  Stoyanov, and Zettlemoyer}]{lewis2019bart}
Mike Lewis, Yinhan Liu, Naman Goyal, Marjan Ghazvininejad, Abdelrahman Mohamed,
  Omer Levy, Ves Stoyanov, and Luke Zettlemoyer. 2019.
\newblock Bart: Denoising sequence-to-sequence pre-training for natural
  language generation, translation, and comprehension.
\newblock \emph{arXiv preprint arXiv:1910.13461}.

\bibitem[{Liu and Lapata(2019)}]{liu2019text}
Yang Liu and Mirella Lapata. 2019.
\newblock Text summarization with pretrained encoders.
\newblock \emph{arXiv preprint arXiv:1908.08345}.

\bibitem[{Maintainers(2020)}]{numpy2020style}
Numpydoc Maintainers. 2020.
\newblock \href
  {https://numpydoc.readthedocs.io/en/latest/format.html#docstring-standard}
  {Numpydoc docstring guide}.
\newblock Technical report.

\bibitem[{Moreno et~al.(2013)Moreno, Aponte, Sridhara, Marcus, Pollock, and
  Vijay-Shanker}]{moreno2013automatic}
Laura Moreno, Jairo Aponte, Giriprasad Sridhara, Andrian Marcus, Lori Pollock,
  and K~Vijay-Shanker. 2013.
\newblock Automatic generation of natural language summaries for java classes.
\newblock In \emph{2013 21st International Conference on Program Comprehension
  (ICPC)}, pages 23--32. IEEE.

\bibitem[{Moreno et~al.(2014)Moreno, Bavota, Di~Penta, Oliveto, Marcus, and
  Canfora}]{moreno2014automatic}
Laura Moreno, Gabriele Bavota, Massimiliano Di~Penta, Rocco Oliveto, Andrian
  Marcus, and Gerardo Canfora. 2014.
\newblock Automatic generation of release notes.
\newblock In \emph{Proceedings of the 22nd ACM SIGSOFT International Symposium
  on Foundations of Software Engineering}, pages 484--495.

\bibitem[{Radford et~al.(2018)Radford, Narasimhan, Salimans, and
  Sutskever}]{radford2018improving}
Alec Radford, Karthik Narasimhan, Tim Salimans, and Ilya Sutskever. 2018.
\newblock Improving language understanding by generative pre-training.
\newblock \emph{URL https://s3-us-west-2. amazonaws.
  com/openai-assets/researchcovers/languageunsupervised/language understanding
  paper. pdf}.

\bibitem[{Raffel et~al.(2019)Raffel, Shazeer, Roberts, Lee, Narang, Matena,
  Zhou, Li, and Liu}]{raffel2019exploring}
Colin Raffel, Noam Shazeer, Adam Roberts, Katherine Lee, Sharan Narang, Michael
  Matena, Yanqi Zhou, Wei Li, and Peter~J Liu. 2019.
\newblock Exploring the limits of transfer learning with a unified text-to-text
  transformer.
\newblock \emph{arXiv preprint arXiv:1910.10683}.

\bibitem[{Raychev et~al.(2014)Raychev, Vechev, and Yahav}]{raychev2014code}
Veselin Raychev, Martin Vechev, and Eran Yahav. 2014.
\newblock Code completion with statistical language models.
\newblock In \emph{Proceedings of the 35th ACM SIGPLAN Conference on
  Programming Language Design and Implementation}, pages 419--428.

\bibitem[{Scalabrino et~al.(2017)Scalabrino, Bavota, Vendome,
  Linares-V{\'a}squez, Poshyvanyk, and Oliveto}]{scalabrino2017automatically}
Simone Scalabrino, Gabriele Bavota, Christopher Vendome, Mario
  Linares-V{\'a}squez, Denys Poshyvanyk, and Rocco Oliveto. 2017.
\newblock Automatically assessing code understandability: How far are we?
\newblock In \emph{2017 32nd IEEE/ACM International Conference on Automated
  Software Engineering (ASE)}, pages 417--427. IEEE.

\bibitem[{Svyatkovskiy et~al.(2020)Svyatkovskiy, Deng, Fu, and
  Sundaresan}]{svyatkovskiy2020intellicode}
Alexey Svyatkovskiy, Shao~Kun Deng, Shengyu Fu, and Neel Sundaresan. 2020.
\newblock Intellicode compose: Code generation using transformer.
\newblock \emph{arXiv preprint arXiv:2005.08025}.

\bibitem[{Svyatkovskiy et~al.(2019)Svyatkovskiy, Zhao, Fu, and
  Sundaresan}]{svyatkovskiy2019pythia}
Alexey Svyatkovskiy, Ying Zhao, Shengyu Fu, and Neel Sundaresan. 2019.
\newblock Pythia: Ai-assisted code completion system.
\newblock In \emph{Proceedings of the 25th ACM SIGKDD International Conference
  on Knowledge Discovery \& Data Mining}, pages 2727--2735.

\bibitem[{Vaswani et~al.(2017)Vaswani, Shazeer, Parmar, Uszkoreit, Jones,
  Gomez, Kaiser, and Polosukhin}]{vaswani2017attention}
Ashish Vaswani, Noam Shazeer, Niki Parmar, Jakob Uszkoreit, Llion Jones,
  Aidan~N Gomez, {\L}ukasz Kaiser, and Illia Polosukhin. 2017.
\newblock Attention is all you need.
\newblock In \emph{Advances in neural information processing systems}, pages
  5998--6008.

\bibitem[{Voita et~al.(2019)Voita, Sennrich, and Titov}]{voita2019bottom}
Elena Voita, Rico Sennrich, and Ivan Titov. 2019.
\newblock The bottom-up evolution of representations in the transformer: A
  study with machine translation and language modeling objectives.
\newblock \emph{arXiv preprint arXiv:1909.01380}.

\bibitem[{Wan et~al.(2018)Wan, Zhao, Yang, Xu, Ying, Wu, and
  Yu}]{wan2018improving}
Yao Wan, Zhou Zhao, Min Yang, Guandong Xu, Haochao Ying, Jian Wu, and Philip~S
  Yu. 2018.
\newblock Improving automatic source code summarization via deep reinforcement
  learning.
\newblock In \emph{Proceedings of the 33rd ACM/IEEE International Conference on
  Automated Software Engineering}, pages 397--407.

\bibitem[{Wang et~al.(2020)Wang, Zhang, Zeng, and Xu}]{wang2020trans}
Wenhua Wang, Yuqun Zhang, Zhengran Zeng, and Guandong Xu. 2020.
\newblock Trans\^{} 3: A transformer-based framework for unifying code
  summarization and code search.
\newblock \emph{arXiv preprint arXiv:2003.03238}.

\bibitem[{Wolf et~al.(2019)Wolf, Debut, Sanh, Chaumond, Delangue, Moi, Cistac,
  Rault, Louf, Funtowicz, and Brew}]{Wolf2019HuggingFacesTS}
Thomas Wolf, Lysandre Debut, Victor Sanh, Julien Chaumond, Clement Delangue,
  Anthony Moi, Pierric Cistac, Tim Rault, R'emi Louf, Morgan Funtowicz, and
  Jamie Brew. 2019.
\newblock Huggingface's transformers: State-of-the-art natural language
  processing.
\newblock \emph{ArXiv}, abs/1910.03771.

\bibitem[{Yin and Neubig(2017)}]{yin-neubig-2017-syntactic}
Pengcheng Yin and Graham Neubig. 2017.
\newblock \href {https://doi.org/10.18653/v1/P17-1041} {A syntactic neural
  model for general-purpose code generation}.
\newblock In \emph{Proceedings of the 55th Annual Meeting of the Association
  for Computational Linguistics (Volume 1: Long Papers)}, pages 440--450,
  Vancouver, Canada. Association for Computational Linguistics.

\bibitem[{Zhai et~al.(2019)Zhai, Xu, Shi, Pan, Ma, Xu, Zhang, Tan, and
  Zhang}]{zhai2019cpc}
Juan Zhai, Xiangzhe Xu, Yu~Shi, Minxue Pan, Shiqing Ma, Lei Xu, Weifeng Zhang,
  Lin Tan, and Xiangyu Zhang. 2019.
\newblock Cpc: automatically classifying and propagating natural language
  comments via program analysis.

\end{thebibliography}

\end{document}